\documentclass[a4paper,conference]{IEEEtran}
\pdfoutput=1
\usepackage{amsmath}
\usepackage{amssymb}
\usepackage{mathtools}
\usepackage{bm}  
\usepackage{xfrac}  

\usepackage{booktabs}
\usepackage{multirow}
\usepackage{colortbl,dcolumn}

\usepackage{algpseudocode}  
\makeatletter
\NewDocumentCommand{\LeftComment}{s m}{%
  \Statex \IfBooleanF{#1}{\hspace*{\ALG@thistlm}}\(\triangleright\) #2}
\makeatother

\graphicspath{{figures/}}

\ifCLASSOPTIONcompsoc
    \usepackage[caption=false,font=normalsize,labelfont=sf,textfont=sf]{subfig}
\else
    \usepackage[caption=false,font=footnotesize]{subfig}
\fi

\usepackage{microtype}

\usepackage[breaklinks=true,
            colorlinks,
            bookmarks=false,
            backref=page
            ]{hyperref}
\usepackage[capitalize]{cleveref}  
\crefname{table}{Table}{Tables}
\crefname{figure}{Figure}{Figures}
\crefname{equation}{}{}
\crefrangeformat{equation}{(#3#1#4) -- (#5#2#6)}  
\usepackage{cite}  

\usepackage{siunitx}
\usepackage{xspace}  
\usepackage{xcolor}
\usepackage[inline]{enumitem}
\usepackage{lipsum}  

\input{settings/commands}
\usepackage[acronym]{glossaries}
\glsdisablehyper

\newcommand*\myglsentry[1]{%
  \protect\ifglsused{#1}{%
    \glsentryshort{#1}%
  }{%
    \glsentrylong{#1}%
  }%
}

\newacronym{dbscan}{DBSCAN}{Density-Based Spatial Clustering of Applications with Noise}
\newacronym{dir}{DIR}{Detection and Identification Rate}
\newacronym{dnn}{DNN}{deep neural network}
\newacronym{evm}{EVM}{Extreme Value Machine}
\newacronym{ev}{EV}{extreme vector}
\newacronym{ievm}{iEVM}{incremental \myglsentry{evm}}
\newacronym{evt}{EVT}{Extreme Value Theory}
\newacronym{far}{FAR}{False Alarm Rate}
\newacronym{gmc}{GMC}{Generalized Maximum Coverage}
\newacronym{lda}{LDA}{Linear Discriminant Analysis}
\newacronym{lfw}{LFW}{Labeled Faces in the Wild}
\newacronym{mpi}{MPI}{Maximum Probabiltiy of Inclusion}
\newacronym{mst}{MST}{Maximum Spanning Tree}
\newacronym{ncm}{NCM}{Nearest Class Mean}
\newacronym{nno}{NNO}{Nearest Non-Outlier} 
\newacronym{ood}{OOD}{out-of-distribution}
\newacronym{rbf}{RBF}{radial basis function}
\newacronym{osr}{OSR}{open set recognition}
\newacronym{owr}{OWR}{open world recognition}

\newacronym{nn}{NN}{Nearest Neighbor}
\newacronym{knn}{$k$-NN}{$k$-\myglsentry{nn}}
\newacronym{osnn}{OSNN}{\OpenSet \myglsentry{nn}}
\newacronym{osnno}{OSNN}{\OpenSet \myglsentry{nn}}
\newacronym{tnn}{TNN}{Thresholded \myglsentry{nn}}

\newacronym{svm}{SVM}{Support Vector Machine}
\newacronym{wsvm}{W-SVM}{Weibull \myglsentry{svm}}
\newacronym{pisvm}{$P_I$-SVM}{Probability of Inclusion \myglsentry{svm}}

\usepackage{tikz}
\usepackage{tikzscalefix}
\usepackage{pgfplots}
\usepgfplotslibrary{groupplots}
\usetikzlibrary{shapes, external}
\newcommand{\tikznewsample}{%
    \begin{tikzpicture}[baseline=-1mm]
    \node[star, draw=black!60!yellow, fill=yellow, scale=.75] (s) at (0,0) {};
    \end{tikzpicture}%
}

\newcommand{\tikzextremevec}{%
    \begin{tikzpicture}[baseline=-1mm]
    \node[regular polygon, regular polygon sides=7, fill=blue, scale=.75] (c) at (0,0) {};
    \end{tikzpicture}%
}

\newcommand{\tikznormalsample}{%
    \begin{tikzpicture}[baseline=-1mm]
    \node[circle, draw=black!60!green, fill=green, scale=.75] (s) at (0,0) {};
    \end{tikzpicture}%
}

\definecolor{color-osnn}{rgb}{.255,.714,.769}  
\definecolor{color-evm}{rgb}{0.533,.337,.855}  
\definecolor{color-c-evm}{rgb}{.192,.639,.329}  
\definecolor{color-extra1}{rgb}{0., 0., .55}  
\definecolor{color-tnn}{rgb}{1.,.55,0.}  

\pgfplotsset{
	compat=1.16,
}

\pgfplotsset{my-basestyle/.style={
	label style={font=\scriptsize,inner sep=1pt},
	tick label style={font=\scriptsize},
    no markers,
	grid=both,
    every axis plot/.append style={line width=1.4pt},
    enlarge x limits={abs=.5em},
    enlarge y limits={abs=.5em},
	legend style={font=\scriptsize},
    legend cell align=left,
}}

\pgfplotsset{style-tnn/.style={
    solid, 
    color=color-tnn,
}}

\pgfplotsset{style-osnn/.style={
    solid, 
    color=color-osnn,
}}

\pgfplotsset{style-evm/.style={
    solid, 
    color=color-evm,
}}

\pgfplotsset{style-evm-sc/.style={
    style-evm,
    dotted,
}}

\pgfplotsset{style-ievm/.style={
    style-evm,
    dashed,
}}

\pgfplotsset{style-cevm/.style={
    solid, 
    color=color-c-evm,
}}

\pgfplotsset{style-cievm/.style={
    style-cevm,
    dashed,
}}

\pgfplotsset{style-extra1/.style={
    solid, 
    color=color-extra1,
}}

\makeatletter
\def\ps@IEEEtitlepagestyle{%
  \def\@oddfoot{\mycopyrightnotice}%
  \def\@evenfoot{}%
}
\def\mycopyrightnotice{%
  {\footnotesize%
    \begin{minipage}{\textwidth}%
    Accepted at ICPR 2022.\\
    \textcopyright\ 2022 IEEE.  Personal use of this material is permitted.  Permission from IEEE must be obtained for all other uses, in any current or future media, including reprinting/republishing this material for advertising or promotional purposes, creating new collective works, for resale or redistribution to servers or lists, or reuse of any copyrighted component of this work in other works.
    \end{minipage}}%
  \gdef\mycopyrightnotice{}%
}
\@ifundefined{showcaptionsetup}{}{%
 \PassOptionsToPackage{caption=false}{subfig}}
\usepackage{subfig}
\makeatother

\begin{document}

\title{Exploring the \OpenWorld Using\\Incremental Extreme Value Machines}

\author{\IEEEauthorblockN{Tobias Koch\IEEEauthorrefmark{1}, Felix Liebezeit\IEEEauthorrefmark{1},
Christian Riess\IEEEauthorrefmark{2}, Vincent Christlein\IEEEauthorrefmark{2}, and Thomas Köhler\IEEEauthorrefmark{1}}
    \IEEEauthorblockA{\IEEEauthorrefmark{1}e.solutions GmbH, Erlangen, Germany\\Email: tobias.koch@esolutions.de}
    \IEEEauthorblockA{\IEEEauthorrefmark{2}Friedrich-Alexander-Universität Erlangen-Nürnberg, Germany}}

\maketitle

\begin{abstract}
Dynamic environments require adaptive applications.
One particular machine learning problem in dynamic environments is \openworld recognition.
It characterizes a continuously changing domain where only some classes are seen in one batch of the training data and such batches can only be learned incrementally. 
\Openworld recognition is a demanding task that is, to the best of our knowledge, addressed by only a few methods.
This work introduces a modification of the widely known \gls{evm} to enable \openworld recognition.
Our proposed method extends the EVM with a partial model fitting function by neglecting unaffected space during an update.
This reduces the training time by a factor of $28$.
In addition, we provide a modified model reduction using weighted maximum $K$-\setcover to strictly bound the model complexity and reduce the computational effort by a factor of \num{3.5} from \SIrange{2.1}{0.6}{s}.
In our experiments, we rigorously evaluate \emph{openness} with two novel evaluation protocols.
The proposed method achieves superior accuracy of about \SI{12}{\percent} and computational efficiency in the tasks of image classification and face recognition. 
\end{abstract}
\glsresetall

\section{Introduction}\label{sec:introduction}

Traditionally, machine learning treats the world as \textit{closed} and \textit{static} space. In particular for classification, domain data is assumed to comprise pre-defined classes with stationary class-conditional distributions. Also datasets to fit models before deploying them shall be available in a single chunk. Practitioners develop such models under controlled lab conditions, where they nowadays rely on tremendous computational resources.

This scarcely applies to many \realworld applications as the world is an \textit{open} space in many facets. For instance, classifiers might be confronted with classes unseen during training. Also distributions of pre-trained classes might be non-stationary or models shall learn novel classes within operation mode. These aspects often occur simultaneously like in image classification, where unknown image categories should be distinguished from known ones showing \textit{concept drifts} (\eg, captured new data with different cameras). It is also in the very nature of biometric systems like face or writer identification that are confronted with known subjects having concept drifts (\eg., due to aging or environmental changes), novel subjects to enroll, and unknown subjects. There is also a steady quest for making the respective algorithms computationally efficient to be applicable on edge devices with limited resources.

\Gls{owr} as formalized by Bendale and Boult~\cite{bendale2015openworld} addresses such constraints and includes three subtasks.
\begin{enumerate*}
    \item \emph{Recognize} new samples either as a \emph{known} or \emph{unknown}.
    \item \emph{Label} new samples either by approving the recognition or defining a new known class.
    \item \emph{Adapt} the current model by exploiting updated labels.
\end{enumerate*}

The recognition subtask poses an independent research area termed \gls{osr} \cite{scheirer2012openset} and received a lot of interest in applications like face recognition~\cite{gunther2017opensetface}, novelty and intrusion detection~\cite{bendale2016opensetnn, henrydoss2017ievm, prijatelj2021novelty}, and forensics~\cite{lorch2020jpeg, maier2020bnn, lorch2021gps}. 
Currently \gls{evm} models as proposed by Rudd \etal~\cite{rudd2017evm} are state of the art in \gls{osr}. 
\Glspl{evm} predict unnormalized class-wise probabilities for query samples to be included in the respective known classes. 
Model fitting depends on class negatives, \ie, it adapts well to imbalanced data, which is a common problem in incremental learning~\cite{ditzler2010learn++unb, wu2019largeir}. 
However, fitting and prediction scale badly for large datasets making their use on resource limited platforms difficult.

Model adaptability can be achieved by cyclic retraining. However, this model-agnostic approach is computationally inefficient and all data needs to be organized in a single chunk. \textit{Incremental learning} aims at doing adaptions effectively and efficiently by batch-wise or sample-wise incorporation of novel data. This needs to handle different challenges: On the one hand, data undergoes concept drifts that shall be learned. On the other hand, the stability-plasticity dilemma~\cite{carpenter1987stabilityplasticity} could either lead to maximum predictive power on previously learned classes (\ie, high stability) or on novel classes (\ie, high plasticity). A good tradeoff between both border cases is desired for well-generalizing models.
Although there are several incremental formulations of popular classifiers~\cite{bifet2009adaptivedt, cauwenberghs2001isvc} or deep learning architectures~\cite{rebuffi2017icarl, castro2018endir, wu2019largeir}, these approaches assume closed sets of known classes in their prediction phase. In principle, probabilistic models like the \gls{evm} can handle batch-wise data but their actual behaviour in incremental learning under an \openworld regime is still widely unexplored. In this paper, we show that simple ad-hoc applications of existing \gls{evm} approaches in \gls{owr} lead to suboptimal stability-plasticity tradeoffs.

The contribution of this paper can be summarized as follows:
\begin{enumerate*}
    \item A partial model fitting algorithm that prevents costly Weibull estimations by neglecting unaffected space during an update. This reduces the incremental training time by a factor of $28$.
    \item A model reduction technique using weighted maximum $K$-\setcover providing fixed size model complexities, which is fundamental for memory constrained systems.
    This approach is up to $4\times$ faster than existing methods and achieves higher recognition rates of about \SI{12}{\percent}.
    \item Two novel \openworld protocols that can be adapted to vary the task complexity in terms of openness.
    \item The framework is evaluated on these protocols with varying difficulty and dimensional complexity for applications such as image classification~and~face~recognition.
\end{enumerate*}%

\section{Related Work}\label{sec:related-work}
\subsubsection{Incremental Learning} 
Popular classifiers such as \glspl{svm}, decision trees, linear discriminant analysis, and ensemble techniques are modified to allow efficient model adaptations~\cite{domingos2000hoeffdingtrees, cauwenberghs2001isvc, polikar2001learnpp, kim2007incrementallda, bifet2009adaptivedt}.
Curriculum and self-paced learning are concepts to sequentially incorporate samples into a model in a meaningful order~\cite{bengio2009curriculum, kumar2010spl, lin2017activeselfpaced}.
iCaRL~\cite{rebuffi2017icarl} and EEIL~\cite{castro2018endir} use distillation or bias correction~\cite{wu2019largeir} to counter catastrophic forgetting.
Zhang \etal~\cite{zhang2021fewshotil} proposed a pseudo incremental learning paradigm by decoupling the feature and classification learning stages.
However, the adaptation of underlying \glspl{dnn} on embedded hardware, as required in many open world applications~\cite{bendale2015openworld}, is far from being efficient.
Additionally, these incremental strategies are not designed for \gls{osr}.

\subsubsection{\OpenSet Recognition} 
Early approaches~\cite{tax2008growing,bartlett2008classification,grandvalet2008svm,cevikalp2012efficient} define threshold-based unknown detection rules for closed-set classifier outputs.
More recent methods focus on the \gls{evt} to consider negative class samples for the estimation of rejection probabilities.
Scheirer \etal~\cite{scheirer2014wsvm} developed the \gls{wsvm} that combines a one-class and a binary \gls{svm}, where decision scores are calibrated via Weibull distributions. Jain \etal~\cite{jain2014pisvm} proposed the \gls{pisvm} to calibrate the outputs of a \acrshort{rbf} \gls{svm} to unnormalized posterior probabilities.
The related OpenMax~\cite{bendale2016opensetnn} calibration is used for class activations of \glspl{dnn} to model the probability of samples being unknown.
Unfortunately, such re-calibrations do not support incremental learning off-the-shelf.
Also GANs allow to sharpen \openset models with adversarial samples~\cite{ge2017genopenmax, neal2018counterfactual, kong2021opengan, yue2021counterfactualzf}.
Recent novelty detection approaches focus on the uncertainty expressiveness of classifiers that can be used to perform novelty or unknown detection, such as Bayesian neural networks~\cite{blundell2015bnn}, Bayesian logistic regression~\cite{lorch2020jpeg}, and Gaussian processes~\cite{lorch2021gps}.
While these methods commonly require multiple computationally demanding Monte Carlo draws to calculate the predictive uncertainty,
Sun \etal~\cite{sun2021react} propose a non-incremental post hoc approach to handle model overconfidence.

\subsubsection{Open World Recognition}
\Gls{nn} based classifiers are \openworld capable, as they typically have no actual training step.
The \gls{osnn}~\cite{junior2017osnn} defines the \openspace via a threshold on the ratio of similarity scores of the two most similar classes.
Bendale and Boult~\cite{bendale2015openworld} derived the \gls{nno} algorithm from the \gls{ncm} classifier~\cite{mensink2013ncm, ristin2014incm}.
\gls{nno} rejects samples that are not in the range of any class center where the distance depends on a learned Mahalanobis distance.
However, these approaches are purely distance-based and do not take distributional information into account.
Joseph \etal~\cite{joseph2021ore} proposed an \openworld object detection method that includes fine-tuning of a \gls{dnn} which is typically too costly for embedded hardware.
To overcome the limitations of \glspl{nn}, Rudd \etal~\cite{rudd2017evm} introduced the \gls{evm} that defines sample-wise inclusion probabilities in dependence of their neighborhood of other classes.
Since this approach is based on a \gls{nn}-like data structure, they propose a model reduction technique to keep the most relevant data points, similar to the support vectors of \glspl{svm}, to reduce the memory footprint.
The \gls{evm} has achieved state-of-the-art results in intrusion detection~\cite{henrydoss2017ievm} and \openset face recognition\cite{gunther2017opensetface}.
The C-EVM~\cite{henrydoss2020cevm} performs a clustering prior to the actual \gls{evm} fitting to reduce the dataset size.
These centroids are then used to fit the \gls{evm}.
However, the clustering does not ensure a reduced model size and especially for small batches, it can cause computational overhead.
In contrast, our proposed method adequately detects unaffected space in incremental updates and prevents redundant parameter estimations.
Additionally, we provide a computationally more efficient model reduction using weighted maximum $K$-set cover, that reduces the model size to a fixed user-set value.%

\section{Background: Extreme Value Theory}\label{sec:background}
The \gls{evm} estimates per-sample probabilities of inclusions. 
Let $\Bx_i$ be a feature vector of class $y_i$ referred to as an anchor sample. Given $(\Bx_i, y_i)$, we select the $\tau$ nearest negative neighbors $\Bx_j$, $j = 1, \ldots, \tau$ from different classes $y_j \neq y_i$ according to a distance $d(\Bx_i, \Bx_j)$, where $\tau$ denotes a \tailsize.
The inclusion probability of a sample $\Bx$ for class $y_i$ is given by the cumulative Weibull distribution:
\begin{equation}
    \Psi_i(\Bx) = \Psi(\Bx; \theta_i) = \exp{\left(- \left( \frac{d(\Bx_i, \Bx)}{\lambda_i} \right)^{\kappa_i} \right)} \enspace \tc
\end{equation}
where $\theta_i = \{\kappa_i, \lambda_i\}$ denotes the Weibull parameters, $\kappa_i$ is the \emph{shape}, and $\lambda_i$ is the \emph{scale} associated with $\Bx_i$.
Given labeled training data $\MN = \left\{ (\Bx_1, y_1), \ldots, (\Bx_N, y_N) \right\}$, each feature vector $\Bx_i$ with class label $y_i$ becomes an anchor. Fitting the underlying EVM aims at sample-wise estimating their~$\theta$.
A query sample $\Bx$ is assigned to class $y_i$ with maximum probability $\max_{i \in N} \Psi_i(\Bx)$. 
This probability shall reach a threshold $\delta$ to distinguish knowns and unknowns according to:
\begin{equation}
    y = 
    \begin{cases}
        y_i & \text{if } \max_{i \in N} \Psi_i(\Bx) \geq \delta \enspace \tc \\
        \text{``unknown''} & \text{otherwise} \enspace \td
    \end{cases}
\end{equation}

A baseline approach keeps all $\theta_i$, which is expensive in terms of prediction time complexity and memory footprint. 
Rudd~\etal~\cite{rudd2017evm} proposed a model reduction such that only informative $\theta_i$, \emph{\glspl{ev}}, are kept since samples within the same class might be redundant.
It can be expressed as \setcover problem~\cite{karp1972setcover} to find a minimum number of samples that \emph{cover} all other samples.
Redundancies are determined by inclusion probabilities $\Psi_i(\Bx_j)$ within $N_c$ samples of a class $c$ ($y_i = y_j \, \forall i,j \in \{1,\ldots,N_c\}$).
A sample $\Bx_j$ is discarded if it is covered by $\theta_i$, \ie, $\Psi_i(\Bx_j) \geq \zeta$, where $\zeta$ denotes the coverage threshold. 
This can be formulated as the minimization~problem:
\begin{align}
    \text{minimize} \enspace \sum_{i=1}^{N_c} I(\theta_i) \label{eq:rudd-optimize} 
    \enspace \text{subject to} \enspace I(\theta_i) \Psi_i(\Bx_j) \geq \zeta \enspace \tc
\end{align}
where the indicator function $I(\theta_i)$ is given by:
\begin{equation}\label{eq:rudd-indicator-function}
    I(\theta_i) = 
    \begin{cases}
        1 & \text{if any } \Psi_i(\Bx_j) \geq \zeta \quad \forall j \in N_c \, \enspace \tc \\
        0 & \text{otherwise}\enspace \td
    \end{cases}
\end{equation}

Rudd~\etal~\cite{rudd2017evm} determines approximate solutions in $\mathcal{O}(N_c^2)$ using greedy iterations, where in each iteration samples that cover most other samples are selected.
This approach does not constrain the amount of \glspl{ev}, which might be necessary for memory limited systems. 
To this end, bisection to determine a suitable $\zeta$ \emph{per class} can be performed.

\section{Incremental Extreme Value Learning}\label{sec:incremental-learning}
During online learning new data points arise and may interfere with the current \glspl{ev}' Weibull distribution estimates.

\subsubsection{Incremental Learning Framework}\label{ssec:incremental-framework}
\gls{evm} learning involves two subtasks: 
\begin{enumerate*}
    \item \emph{Model fitting} to adapt the model to new data and
    \item \emph{model reduction} that bounds the model's computational complexity and required resources.
\end{enumerate*}
In \gls{owr}, both steps need to handle training data arriving batch-wise over consecutive epochs.
We perform incremental learning over epochs using new arriving training batches $\MN^t$, where $t$ denotes the epoch index.
For an incremental formulation, let $\Theta_E^t = \{\theta_1^t, \ldots, \theta_E^t\}$ be a model of $E$ \glspl{ev} determined either at the previous epoch or learned from scratch at the first epoch.
The fit function incorporates the new batch $\MN^t$ to the current model $\Theta_E^t$ to obtain a new intermediate model $\Theta^{t+1}$.
The reduction squashes $\Theta^{t+1}$ according to a given budget by selecting most informative \glspl{ev} considering both previous and new samples. This yields the consolidated model $\Theta_{E}^{t+1} \subseteq \Theta^{t+1}$.
Our framework alternates the fit and reduction function efficiently per epoch.

\subsubsection{Partial Model Fitting}
\label{ssec:partial-model-fit}
\begin{figure}[tb]
    \centering
    \subfloat[No update required.]{\tikzsetnextfilename{incremental-updates1}
\begin{tikzpicture}[
    sample/.style={circle, draw=black!60!green, fill=green, scale=0.65}
    ]
    \node[regular polygon, regular polygon sides=7, fill=blue] (c0) at (0,0) {};
    \node[circle, draw=black!60!blue, fill=blue, fill opacity=0.25, scale=7] (c1) at (0,0) {};

    \node[sample] (s1) at (0.25, 0.22) {};
    \node[sample] (s2) at (0.22, 0.45) {};
    \node[sample] (s3) at (-0.35, 0.5) {};
    \node[sample] (s4) at (-0.7, -0.9) {};
    \draw[thick]  (c0) -- node [auto] (dt) {$d_\tau$} (s4);

    \node[star, draw=black!60!yellow, fill=yellow, scale=0.75] (s5) at (-1.5, -0.85) {};
\end{tikzpicture}\label{sfig:incremental-a}}
    \qquad
    \subfloat[Update required.]{\tikzsetnextfilename{incremental-updates2}
\begin{tikzpicture}[
    sample/.style={circle, draw=black!60!green, fill=green, scale=0.65}
    ]
    \node[regular polygon, regular polygon sides=7, fill=blue] (c0) at (0,0) {};
    \node[circle,draw=black!60!blue, dashed, scale=7, outer sep=0] (c1) at (0,0) {};
    \node[circle,draw=black!60!blue,fill=blue,fill opacity=0.25,scale=5, outer sep=0] (c2) at (0,0) {};

    \node[sample] (s1) at (0.25, 0.22) {};
    \node[sample] (s2) at (0.22, 0.45) {};
    \node[sample] (s3) at (-0.35, 0.5) {};
    \node[sample,fill=lightgray] (s4) at (-0.7, -0.9) {};

    \node[star, draw=black!60!yellow, fill=yellow, scale=0.75] (s5) at (-0.75, -0.3) {};
    \draw[thick]  (c0) -- node [auto] (dt) {$d_\tau$} (s5);

    \draw[->, thick]  (c1.north) -- (c2.north);
    \draw[->, thick]  (c1.north east) -- (c2.north east);
    \draw[->, thick]  (c1.east) -- (c2.east);
    \draw[->, thick]  (c1.south east) -- (c2.south east);
    \draw[->, thick]  (c1.south) -- (c2.south);
    \draw[->, thick]  (c1.south west) -- (c2.south west);
    \draw[->, thick]  (c1.west) -- (c2.west);
    \draw[->, thick]  (c1.north west) -- (c2.north west);
\end{tikzpicture}\label{sfig:incremental-b}}
     \caption{Incremental update illustration with $\tau=4$. The Weibull distribution of the \acrfull{ev}~(\protect\tikzextremevec) is estimated on the $\tau$ nearest samples~(\protect\tikznormalsample). The blueish hypersphere with radius~$d_\tau$ is derived from the farthest sample. The new sample~(\protect\tikznewsample) in \cref{sfig:incremental-a} lies outside the sphere and can be ignored. Once a new sample lies within the sphere, \cf \cref{sfig:incremental-b}, an update is required.}
	\label{fig:incremental}%
\end{figure}
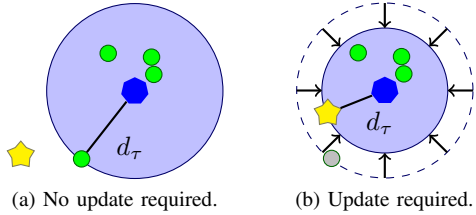%
For model fitting, we process samples in new arriving batches $\MN^t$ independently to incorporate them into the current model $\Theta_E^t$. 
A new sample $\Bx^{t+1}$ might fall into the neighborhood of any \gls{ev}'s feature vector $\Bx_e^t$, which would invalidate the corresponding Weibull parameters in $\theta_e^t$, where $\theta_e^t \in \Theta_E^t$.
A naive approach is to re-estimate a new Weibull distribution for each \gls{ev} including nearest negative neighbor search and tail construction.
We argue that this is highly inefficient since it is most likely that the new sample will not influence all the \glspl{ev}. Thus, most estimates will result in the same Weibull parameters as previously.

We extend the \gls{evm} model by an automatically derivable, \ie, nonuser-set value, namely the \emph{maximum \taildistance}~$d_\tau$, which corresponds to the maximum distance within a tail such that $\theta_e^t = \{\kappa_e^t, \lambda_e^t, d_{\tau,e}^t\}$.
This parameter operates as a threshold and controls the model update.
It can be described by a hypersphere centered at an \gls{ev} with radius $d_{\tau}$ as depicted in \cref{fig:incremental}.
Anytime a sample falls into this hypersphere, we need to shrink it.
To perform partial fits, we need to compute distances between $\Bx^{t+1}$ and all $\Bx_e^t$ and estimate the Weibull parameters for $\Bx^{t+1}$.
Using these distances, we define the update rule for the \gls{ev}:
\begin{equation}
    \theta_e^{t+1} = 
    \begin{cases}
        \text{update}(\theta_e^{t}) & \text{if } d(\Bx_e^t, \Bx^{t+1}) < d_{\tau,e}^t \enspace \tc \\
        \theta_e^{t} & \text{otherwise} \enspace \tc
    \end{cases}
\end{equation}
where $\text{update}(\cdot)$ denotes tail update, re-estimation of Weibull parameters, and storage of new maximum tail distances.
This allows computationally efficient partial fits and leads to exactly the same result as cyclic retraining, as long as no model reduction is carried out.
\begin{table}[tb]
    \caption{Update ratio [\%] of the \acrfullpl{ev} on a subset of MNIST. The lower the ratio the more updates can be skipped.}%
    \label{tab:mnist-update-ratio}%
    \centering
    \setlength{\aboverulesep}{1pt}
    \setlength{\belowrulesep}{1pt}
    \begin{tabular}{ccccc}
    \toprule
      \textbf{Batch Size} & \multicolumn{4}{c}{\textbf{Tail Size $\tau$}} \\
        & 5 & 25 & 100 & 250 \\
      \midrule
        \phantom{00}5   & \phantom{0}\zz{0.56} & \phantom{0}\zz{2.44} & \phantom{0}\zz{8.93} & \zz{20.17} \\
        \phantom{0}25  & \phantom{0}\zz{2.68} & \zz{10.81} & \zz{33.29} & \zz{59.83} \\
        \phantom{0}50  & \phantom{0}\zz{5.15} & \zz{19.26} & \zz{51.55} & \zz{79.40} \\
        100 & \phantom{0}\zz{9.63} & \zz{32.54} & \zz{72.22} & \zz{93.42} \\
        250 & \zz{21.19} & \zz{58.05} & \zz{93.04} & \zz{99.51} \\
        \bottomrule
    \end{tabular}%
\end{table}

In \cref{tab:mnist-update-ratio}, we exemplify the gain of this approach.
We incrementally fit an \gls{evm} on a subset of MNIST and store all samples as \glspl{ev}.
The update ratio determines the fraction of \glspl{ev} that require an update in subsequent epochs.
It follows, the smaller the batches and tail size the less updates are necessary.
The benefit can become very substantial at small batch and tail sizes with an update ratio of only \SI{0.56}{\percent}.
%
\subsubsection{Model Reduction}\label{ssec:model-reduction}
\renewcommand{\thefigure}{2}
\begin{figure*}[tb]
    \setlength{\fboxsep}{0pt}%
    \centering
    \newcommand{\tmpwidth}{.18}
    \subfloat[EVM ($\infty$-SC)~\cite{rudd2017evm}]{\includegraphics[width=\tmpwidth\linewidth]{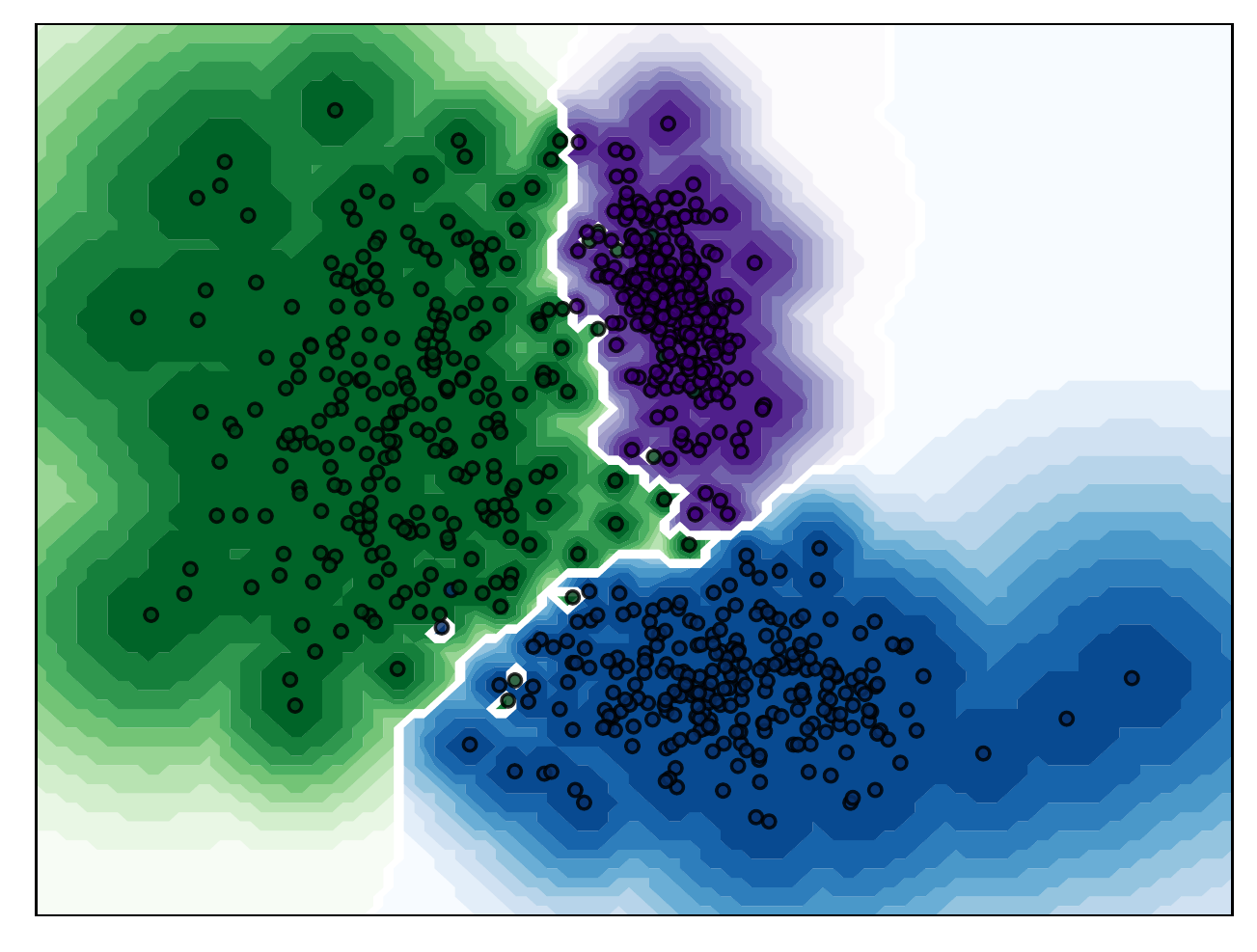}\label{sfig:reduce-none2}}%
    \hfill%
    \subfloat[EVM ($10$-SC)~\cite{rudd2017evm}]{\includegraphics[width=\tmpwidth\linewidth]{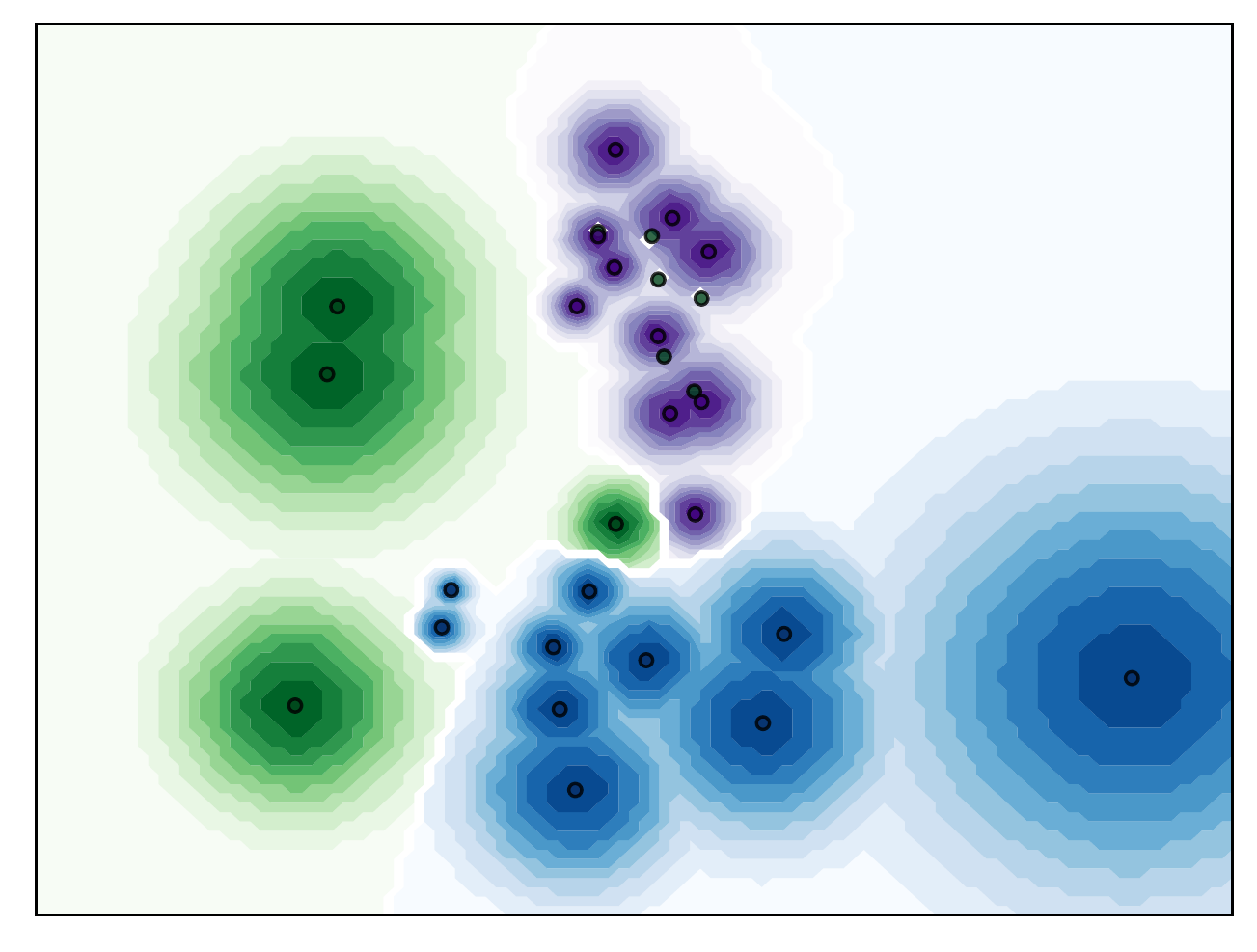}\label{sfig:reduce-orig2}}%
    \hfill%
    \subfloat[iEVM ($10$-wSC)]{\includegraphics[width=\tmpwidth\linewidth]{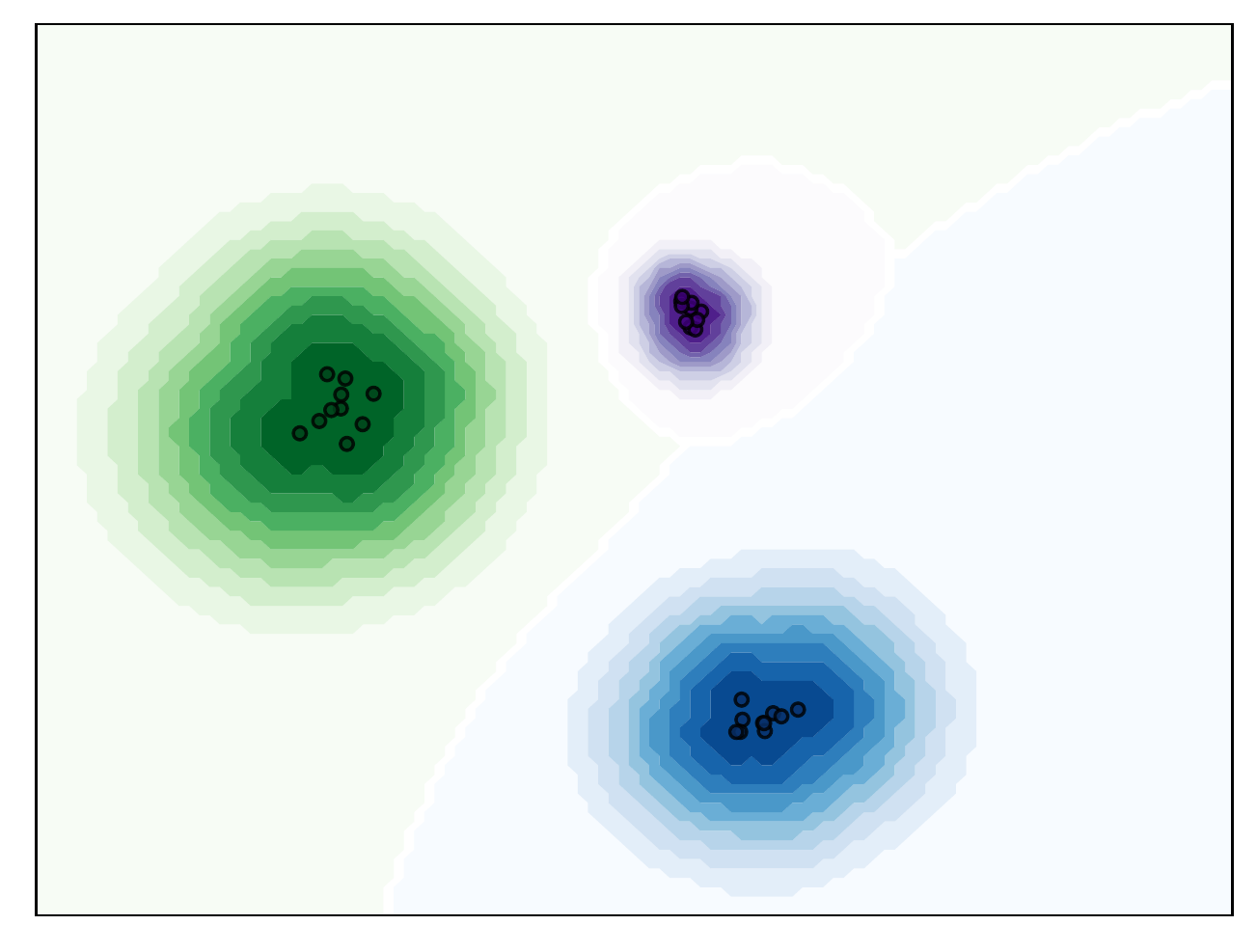}\label{sfig:reduce-ours2}}%
    \hfill%
    \subfloat[C-EVM ($\infty$-SC)~\cite{henrydoss2020cevm}]{\includegraphics[width=\tmpwidth\linewidth]{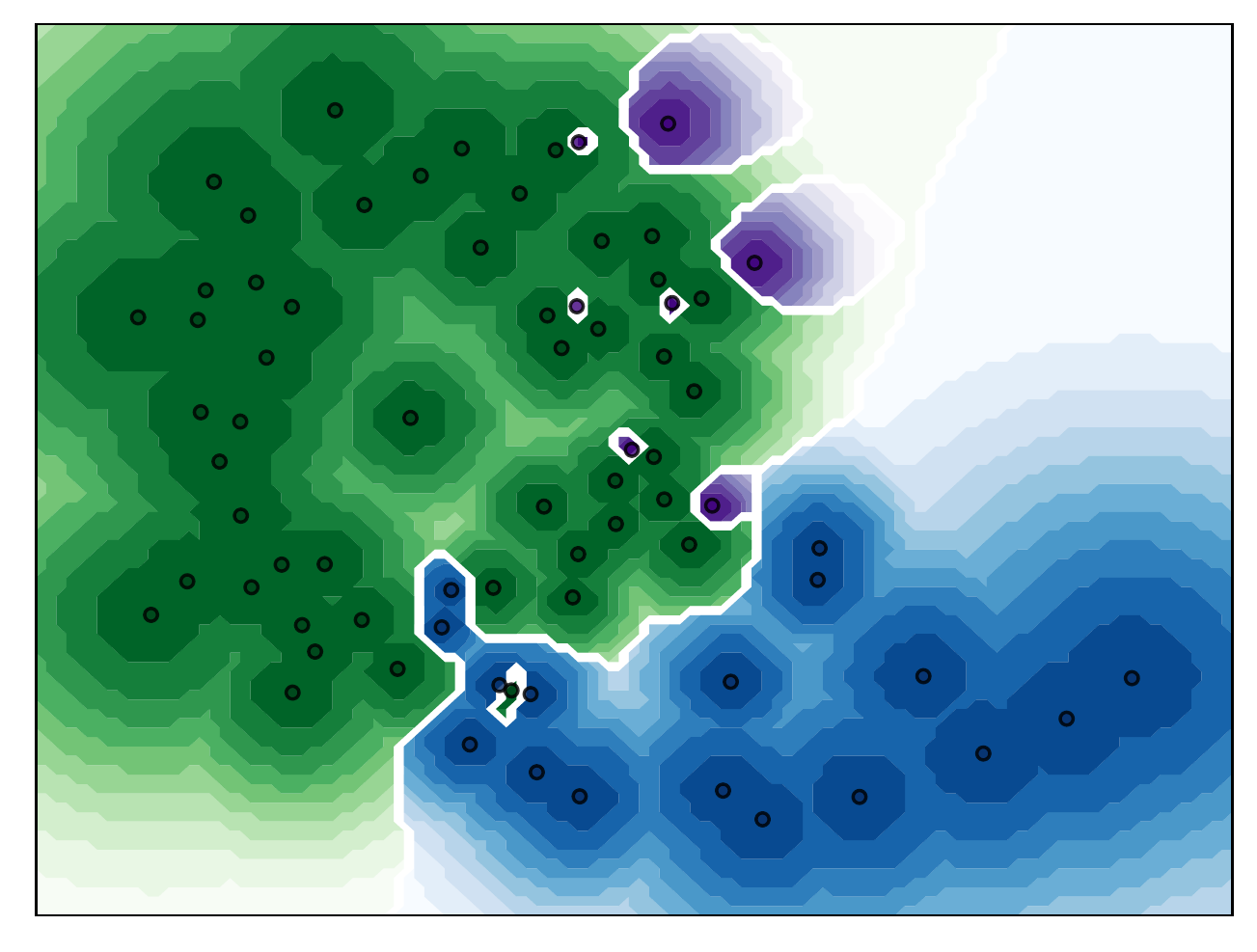}\label{sfig:c-evm2}}%
    \hfill%
    \subfloat[C-iEVM ($10$-wSC)]{\includegraphics[width=\tmpwidth\linewidth]{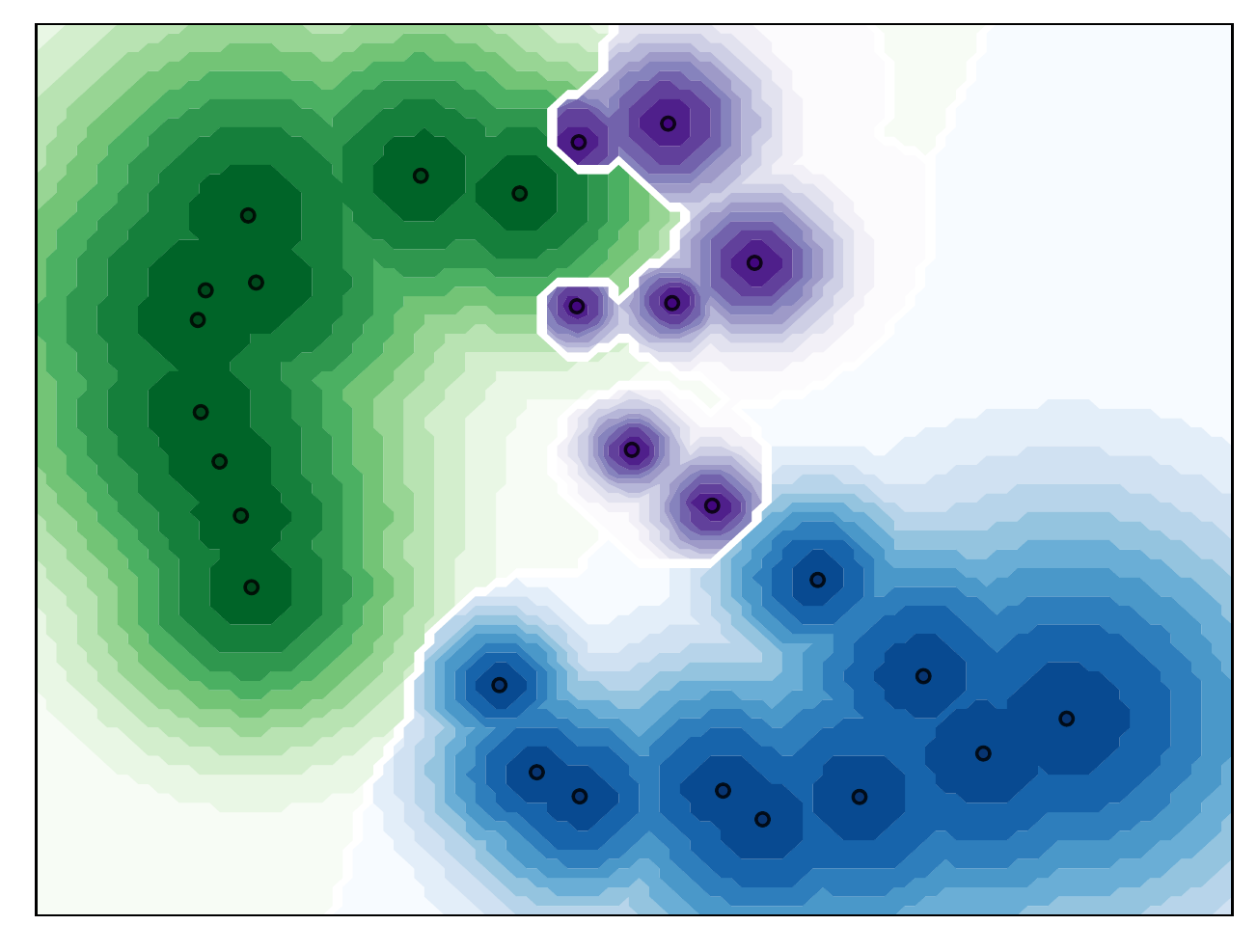}\label{sfig:c-ievm2}}%
    \caption{Decision boundaries of different \gls{evm} reductions on a $3$-class toy dataset. Solid dots correspond to the \acrfullpl{ev} and colored areas belong to the related class where the inclusion probability is visualized via the opacity. 
    In~(a), no reduction is performed, \ie, the \glspl{ev} match the training data. The set cover (SC) reduction and our weighted (wSC) are shown in~(b) 
    and~(c),
    respectively. In~(d), 
    the C-EVM is shown and
    (e)~presents~the~C-EVM with our wSC.}%
    \label{fig:overview-reductions2}%
\end{figure*}%
\renewcommand{\thefigure}{1}  
\begin{figure}[tb]
\renewcommand\figurename{Alg.}  
\vspace{2pt}
\hrule\vspace{2pt}%
\begin{algorithmic}[1]
\Function{reduce}{$\Theta$, $K$}
    \State $\Theta_E \gets \emptyset$
    \For{$k = 1 \text{ \textbf{to} } K$}
        \State idx $\gets \argmax_{i \in |\Theta|} \sum_{j=1}^{|\Theta|} \Psi_i(\Bx_j)$
        \State $\Theta_E \! \text{.insert}(\theta_{\text{idx}})$
        \State $\Theta \text{.remove}(\theta_{\text{idx}})$
    \EndFor
\State \Return $\Theta_E$
\EndFunction
\end{algorithmic}
\vspace{2pt}\hrule\vspace{2pt}
\caption{The proposed weighted maximum $K$-\setcover \gls{evm} model reduction.}%
\label{alg:gmc}%
\end{figure}
In our incremental learning framework, the aim of a class-wise model reduction $g$ is to find a subset $\Theta_{E_c}^t \subseteq \Theta_c^t$ that is budgeted \wrt the number of resulting \glspl{ev}.

\paragraph{Problem Statement}
For the sake of simplicity, let us drop the batch count $t$ and class index $c$, unless it is necessary.
We denote our model reduction by a function $g \colon \Theta \to \Theta_{E}$, where $\Theta_{E}$ underlies the constraint $|\Theta_{E}| \leq K\leq |\Theta| = N$ and $K$ denotes the budget of \glspl{ev} that can be kept for a certain class with $N$ samples.
The intuition behind the design of $g$ is three-fold: 
\begin{enumerate*}
    \item We aim at selecting \glspl{ev} that best cover others according to pair-wise inclusion probabilities.
    \item While pair-wise inclusion probabilities are not symmetric in general, \ie, $\Psi_i(\Bx_j) \neq \Psi_j(\Bx_i)$, high bilateral coverage is common and would introduce a bias towards selecting \glspl{ev} very close to class centroids implying that selecting both $\Psi_i(\Bx_j)$ and $\Psi_j(\Bx_i)$ shall be penalized.
    \item At most $K$ \glspl{ev} shall be selected.
\end{enumerate*}

We propose to formulate $g$ as a weighted maximum $K$-\setcover~\cite{cohen2008gmc}.
Let us define a collection of sets $\MS = \{ \MS_1, \ldots, \MS_N \}$, where $\MS_i = \{ (w_{kl}, w_{lk}) \, | \, 1 \leq k \leq i < l \leq N \}$ models a single~\gls{ev}.
A pair $(w_{kl}, w_{lk}) \in \MS_i$ contains two weights given by the inclusion probabilities $w_{kl} = \Psi_k(\Bx_l)$ and $w_{lk} = \Psi_l(\Bx_k)$.
We determine $g$ according to the integer linear program:
\begin{align}
    \text{maximize} \enspace & \sum_{i=1}^{N} \sum_{j=i+1}^{N} \beta_{ij} \Psi_i(\Bx_j) + \beta_{ji} \Psi_j(\Bx_i) \label{eq:mst-optimization-3} \\*
    \text{subject to} \enspace & \sum_{i=1}^{N} \gamma_i \leq K \enspace \tc \label{eq:mst-constraint-1-3} \\*
    & \beta_{ij} + \beta_{ji} \leq 1 \enspace \tc
    \label{eq:mst-constraint-2-3}
\end{align}
where $\beta_{ij} \in \{0, 1\}$ selects covered elements ($\beta_{ij} = 1 \Leftrightarrow (w_{ij}, w_{ji})~\text{is covered by}~\MS_i$) and $\gamma_i \in \{0, 1\}$ selects kept \glspl{ev} ($\gamma_i = 1 \Leftrightarrow \MS_i~\text{is kept}$).
The objective in \cref{eq:mst-optimization-3} is optimized \wrt $\beta$ and $\gamma$ to maximize the value of the coverage.
The constraint in~\cref{eq:mst-constraint-1-3} limits the amount of \glspl{ev} to the budget~$K$ and \cref{eq:mst-constraint-2-3} penalizes the selections of bilateral coverage.
\paragraph{Incremental Algorithm}
We solve \crefrange{eq:mst-optimization-3}{eq:mst-constraint-2-3} by greedy iterations as depicted in Algorithm~\ref{alg:gmc}.
Our algorithm facilitates incremental learning by reusing intermediate results from the model reduction of the previous epoch, where $\Theta$ denotes the intermediate model of a class from the partial fit function and $K$ is the \gls{ev} budget.
Line $3$ limits the amount of iterations to the desired budget $K$.
In each iteration, we first compute for each sample the sum of inclusion probabilities from all other samples toward it (line $4$).
The element with the highest sum is selected as \gls{ev} (line $5$ - $6$).
In the end, the reduced model $\Theta_E$ is released.
Note that summations in line $4$ do not need to be recomputed in every iteration.
We provide additional implementation details for Algorithm~\ref{alg:gmc} in the supplementary~material.

\paragraph{Relationship to Previous Works~\cite{rudd2017evm, henrydoss2020cevm}} \label{sssec:relationship}
Our weighted maximum $K$-\setcover formulation in \crefrange{eq:mst-optimization-3}{eq:mst-constraint-2-3} generalizes the conventional \setcover model reduction of Rudd~\etal~\cite{rudd2017evm}.
To formulate \cite{rudd2017evm} in our framework, we need to substitute $\Psi_i(\Bx_j)$ and $\Psi_j(\Bx_i)$ in \cref{eq:mst-optimization-3} by $I(\theta_i)$ and $I(\theta_j)$, \ie, the indicator function of \cref{eq:rudd-indicator-function}.
Thus, all samples with coverage probabilities $\geq \zeta$ are weighted uniformly.

The C-EVM~\cite{henrydoss2020cevm} uses class-wise DBSCAN clustering~\cite{ester1996dbscan} and generates centroids from these clusters.
This preconditioning reduces the training set size before the actual EVM is fitted to the centroids.
However, this does not enforce a specific amount of \glspl{ev}.
This is sub-optimal in memory-limited applications, \eg, on edge devices, where fixed model sizes are preferred.

In \cref{fig:overview-reductions2}, we compare different reduction techniques on example data, where $K$-\setcover ($K$-SC) represents Rudd's method~\cite{rudd2017evm} and ($K$-wSC) our weighted $K$-\setcover ($K$-wSC).
It can be observed that $K$-SC leads to scattered decision boundaries and is sensitive to outliers.
Our stand-alone \gls{ievm} is robust against outliers and empowers the \openspace, \cf \cref{sfig:reduce-ours2}.
The C-EVM generates new centroids but does not guarantee a certain amount of \glspl{ev}.
Therefore, we extend it with our $K$-wSC and bilateral coverage regularization.
This selects \glspl{ev} that accurately describe the underlying distributions of known classes.
We argue that both, the iEVM and C-iEVM, perfectly describe different levels of the stability-plasticity tradeoff.
While the iEVM strictly bounds the decision boundaries to dense class centers and leaves more \openspace, it is stable to concept drift.
In contrast, the C-iEVM enables more plasticity as outliers have a high impact on the generated centroids.

The hard thresholding of Rudd~\etal~\cite{rudd2017evm} also comes at the cost of embedding their \setcover into a bisection search to determine a coverage threshold $\zeta$ providing the desired number of \glspl{ev}. Given a bisection termination tolerance of $\epsilon$, the overall model reduction has a time complexity of $\MO(\log(\epsilon^{-1})N^2)$ for a single class comprising $N$ samples. In contrast, our model reduction method avoids thresholding and considers the given budget on the number of \glspl{ev} in a single pass with time complexity $\MO(N^2)$.
This is an important factor for implementations on resource limited devices.%

\renewcommand{\thefigure}{3}
\begin{figure*}[tb]
    \centering
    \vspace{1ex}
    \tikzsetnextfilename{openworld-cifar-group-horizontal}
    \includegraphics[width=.99\linewidth]{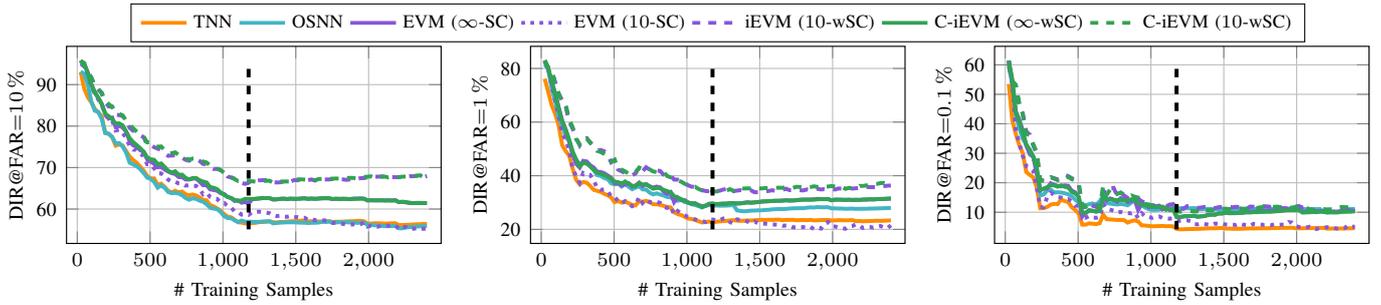}
    \vspace{-1ex}%
    \caption{Averaged results over $3$ runs of the proposed \openworld \protocolA on \cifar{100}. Set cover and our weighted maximum $K$-set cover reduction to $K$ \acrfullpl{ev} are denoted as $K$-SC and $K$-wSC, respectively. The vertical dashed line determines the batch at which the openness remains constant.}
	\label{fig:cifar100-openworld-group}%
\end{figure*}%
\section{\OpenWorld Evaluation Protocols}\label{sec:protocols}
We introduce our two designed \openworld evaluation protocols.
The first protocol describes the very general \realworld online learning environment, where new classes are learned and old classes are updated by new samples.
The second protocol is a specialization of the first one, where subsequent epochs contain only new classes.

\subsubsection{\protocolA}
This protocol reflects the realization of a newly deployed \gls{owr} application.
While others start with a large initial training phase~\cite{bendale2015openworld}, we argue that this is not possible in \realworld scenarios, as the exact environmental conditions, \eg, sensors and lighting, are unknown.
Furthermore, it is an unrealistic assumption to start with a large initial training~phase.

We start with a minimum of $2$ classes and incrementally learn new classes, while incorporating new samples of previous classes.
This introduces two types of concept drifts, termed \emph{direct} and \emph{implicit} concept drift.
Direct concept drift applies to a single changing class, \eg, the aging of a person.
Implicit concept drift determines the mutual impact of neighboring classes competing for transitional feature space.
Here, the occurrence of a new class can have a high impact on previously learned classes as both may share parts of the feature space, \eg, leopards and jaguars.
Implicit concept drift is given whenever an altering class influences the learned concepts of other classes.

\renewcommand{\thefigure}{4}
\begin{figure*}[t]
    \centering
    \vspace{1ex}
    \tikzsetnextfilename{classinc-lfw-macro}
    \includegraphics[width=.99\linewidth]{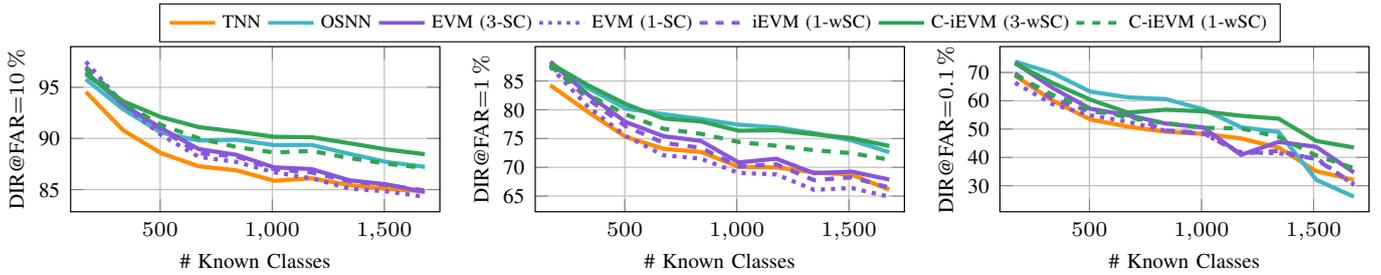}
    \vspace{-1ex}%
    \caption{Averaged results over $3$ runs of \protocolB on LFW. Set cover and our weighted maximum $K$-set cover reduction to $K$ \acrfullpl{ev} are denoted as $K$-SC and $K$-wSC, respectively. Our reduction achieves comparable results \emph{while} reducing the model complexity by factor $4$.}%
	\label{fig:lfw-classinc-macro}%
\end{figure*}%
Our protocol allows the control of its complexity on the basis of an initial \emph{openness}~\cite{scheirer2012openset}.
According to this openness, classes are divided into two disjoint sets of knowns~$\MC_\text{K}$ and unknowns~$\MC_\text{U}$, with $|\cdot|$ denoting the cardinality.
The first epoch contains $2$ classes of $\MC_\text{K}$.
The following epochs comprise a single new class of $\MC_\text{K}$ as well as samples of classes seen in previous epochs.
Hence, all classes in $\MC_\text{K}$ are known at epoch $|\MC_\text{K}| - 1$.
Each learning epoch follows an evaluation on a fixed test set. 
Note that, although the test set is fixed, the amount of unknowns reduces over the epochs.
Thus, the openness decreases from epoch number $1$ to $|\MC_\text{K}| - 1$.
This reduces the complexity of unknown detection while increasing the difficulty for the classification of knowns.
To further investigate the models' incremental adaptability at a steady openness, we continue the epoch-wise training after $|\MC_\text{K}| - 1$ with batches of $\MC_\text{K}$.

\subsubsection{\protocolB}
This protocol specializes the first one for applications with few samples per class.
Due to the limited amount of training samples, we derive a pure class-incremental evaluation, where each epoch contains a certain amount of new classes.
No previously learned classes are directly updated by new samples in subsequent epochs but they are updated implicitly by new occurring classes leading to the previously mentioned implicit concept drift.

We split the classes \wrt a predefined openness into knowns and unknowns.
The unknowns are put in the test set together with a subset of samples for each of the known classes.
The known classes are split into batches where each batch contains all remaining samples of a certain amount of classes.

\subsubsection{Performance Measures}
The \gls{dir} at certain \glspl{far} serves as evaluation metric, which is common in the \openset face recognition~\cite{gunther2017opensetface}. 
The \gls{far} determines the fraction of misclassified unknowns. 
The threshold to receive a certain \gls{far} can be derived from the evaluated dataset.
The \gls{dir} determines the fraction of correctly detected knowns \emph{and} their correct classification.
A high \gls{dir} at low \gls{far} is favorable.

\section{Experiments and Results}\label{sec:experiments}
We evaluate our \gls{ievm} in different \gls{owr} applications.
The \gls{evm}, \gls{osnn}, and \gls{tnn} serve as baselines.
We also extend the C-EVM by our incremental framework, where clustering is applied prior to model fitting.
The method notations are adopted from Section~\ref{sssec:relationship}.
Model reductions are performed at every epoch.

\subsubsection{Image Classification}
The \openworld performance of our approach is evaluated with \protocolA on \cifar{100}~\cite{krizhevsky09cifar}.
This dataset comprises \num{50000} training and \num{10000} test samples of \num{100} classes.
The randomized split into knowns and unknowns is \SI{50}{\percent}, which results in an openness range from \SI{80.2}{\percent} for the first batch to \SI{18.4}{\percent} for batch $49$ and the following ones.
We evaluate $100$ epochs using a batch size of $24$ and benchmark all models on the whole test set after each epoch.
We repeat the protocol $3$ times using different random orders in the creation and processing of batches.

\paragraph{Implementation Details}
For feature extraction, we use EfficientNet-B6~\cite{tan2019efficientnet} pre-trained on ImageNet~\cite{deng2009imagenet} and fine-tuned on a \cifar{100} training split via categorical cross-entropy loss and a bottleneck layer of size $1024$.
All \glspl{evm} use the same parameters: $\tau = 75$ and $\alpha = 0.5$.
For the clustering in the C-EVM and C-iEVM, we adopt the parameters reported in~\cite{henrydoss2020cevm}.
Methods that employ a model reduction reduce the amount of \glspl{ev} to $K = 10$.
We report additional results with alternative parameters in the supplementary material.

\paragraph{Results}
Averaged results of $3$ repetitions of \protocolA are shown in \cref{fig:cifar100-openworld-group}.
We depict the \gls{dir} over the amount of samples at different \glspl{far}.
All \glspl{evm} perform similar for the first \num{250} samples and achieve an initial \gls{dir} of about \SI{95}{\percent} at a \gls{far} of \SI{10}{\percent}. 
In later epochs, our \gls{ievm} and C-iEVM clearly outperform the competing methods for high and medium \glspl{far} (\SI{10}{\percent} and \SI{1}{\percent}), while at very small \gls{far} (\SI{0.1}{\percent}) all methods perform comparably.
However, our methods begin to recover after the openness remains constant.

In the case that the training samples within a class are widely spread, the original \setcover model reduction struggles to find the most important \glspl{ev}.
This leads to a constant decrease in the \gls{dir} even after the openness complexity stays constant.
Similarly, DBSCAN in the C-EVM fails to generate meaningful centroids resulting in almost identical outputs as the baseline \gls{evm}.
We noticed that DBSCAN achieves only average reductions of about \SI{3}{\percent} and the model contains \num{2294}~\glspl{ev} after the last epoch.
Our weighted $K$-\setcover easily selects the most important \glspl{ev} and achieves the best results in the C-iEVM and iEVM while storing only \num{500}~\glspl{ev} ($10$ per class).

The amount of \glspl{ev} does not only influence the memory but also the inference time.
The reduced models take about \SI{2.4}{s} to evaluate the test set while the others require about \SI{14.7}{s} which is a factor of $6$.
Further, our model reduction is, averaged over all epochs, by a factor $4.2$ faster than the conventional one.

\subsubsection{Face Recognition}
To evaluate our method in \openworld face recognition, we apply \protocolB to the \gls{lfw}~\cite{huang2007lfw1, huang2014lfw2} dataset.
We adopt the training and the $O3$~test split of~\cite{gunther2017opensetface}, where the training set consists of \num{2900} samples from \num{1680} unbalanced classes with either $1$ or $3$ images.
We divide this split into $10$ batches with $168$ classes each.
After each epoch the test set is evaluated.
Since the test set is highly unbalanced with \numrange{1}{527} samples per class, we report the \emph{macro} average \gls{dir} at certain \glspl{far}.
This prevents the suppression of misclassified underrepresented classes and is therefore a better representation on the global performance on this dataset.
The protocol is repeated $3$ times.

\paragraph{Implementation Details}
For feature extraction we use the ResNet50, 
pre-trained on MS-Celeb-1M~\cite{guo2016msceleb} and fine-tuned on VGGFace2~\cite{cao2018vggface2}, with an embedding size of \num{128}.
We adopt the \gls{evm} parameters $\tau=75$ and $\alpha=0.5$ from~\cite{gunther2017opensetface}.
Additionally, our methods with model reduction perform the contraction to a single \gls{ev} per class, \ie, $K = 1$.

\paragraph{Results}
We present the averaged \gls{dir} at several \glspl{far} in \cref{fig:lfw-classinc-macro}.
Surprisingly, the \gls{osnn} achieves in this protocol better recognition scores than in the previous one.
The C-EVM and \gls{osnn} perform comparable while the \gls{osnn} looses precision at the lowest FAR (\SI{0.1}{\percent}).
Our C-iEVM and iEVM achieve comparable results \emph{while} reducing the model complexity by a factor of $4$.

The computational efficacy of our incremental framework is presented in \cref{fig:lfw-runtime}.
Here, partial fitting reduces the average training time by a factor of $28$.
In particular, performance gains are substantial at late epochs, where the EVM requires \SI{27}{s} to learn the final classes, while the iEVM takes \SI{0.7}{s}. 
Our model reduction is, averaged over all epochs, by a factor of $3.7$ faster than the conventional set cover approach.
\renewcommand{\thefigure}{5}
\begin{figure}[t]
    \centering
    \tikzsetnextfilename{lfw-runtime}%
    \includegraphics[width=.99\linewidth]{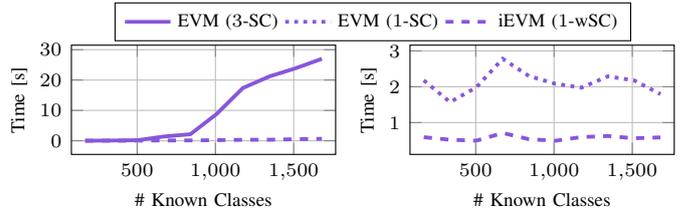}%
    \vspace{-1ex}%
    \caption{Averaged runtime of the training step (left) and model reduction (right) from the evaluation of \protocolB and LFW. Our partial fit reduces the average training time by a factor of $28$. Our model reduction, averaged over all epochs, is faster than the conventional set cover by a factor of $3.7$.}
	\label{fig:lfw-runtime}%
\end{figure}%

\subsubsection{Additional Experiments}
The supplementary material contains additional details about the proposed reduction and the evaluation on an additional dataset~\cite{fiel2017icdar2017} using \protocolB.%

\section{Conclusion}\label{sec:conclusion}
We introduced an incremental leaning framework for the \gls{evm}.
Our partial model fitting neglects unaffected space during an update and prevents costly Weibull estimates.
The proposed weighted maximum $K$-set cover model reduction guarantees a fixed-size model complexity with less computational effort than the conventional set cover approach.
Our reduction leads to dense class centers filtering out outliers.
The proposed modifications outperform the original EVM and the C-EVM on novel \openworld protocols in terms of efficacy and efficiency.
In future work, we will investigate the method on larger datasets to better understand the advantages of our model reduction and put more effort into applications with harsh constraints on low \acrlongpl{far}.%

\clearpage 
\bibliographystyle{IEEEtran}
\bibliography{icpr.bib}
\clearpage
\appendix
\subsection{Algorithm Details}
Algorithm~\ref{alg:gmc2} provides additional details of the proposed weighted maximum $K$-set cover model reduction for the \gls{evm}.
Recall that this is a class-wise reduction technique. 
Thus, the amount of \glspl{ev} in a single class is denoted as $E$. The amount of samples within a batch of this class is denoted $N$. 

The summations of the inclusion probabilities for each \gls{ev} are given in $\Bp$.
The \gls{evm} model $\Theta_E^t$ represents the \glspl{ev} of the previous epoch, $\Theta_N^{t+1}$ the estimated Weibull parameters of the current data batch, and $K$ determines the \gls{ev} budget. 
The reduction comprises four steps: 
\begin{enumerate}
    \item Updating the inclusion probability sums of the old \glspl{ev} \wrt the new batch (line \numrange{2}{4}).
    \item Sum up the inclusion probabilities of the new samples \wrt each other (line \numrange{6}{9}). 
    This step has a time complexity of $\MO(N \cdot (E + N))$ which is $\MO(N^2)$ for large batches (\ie, $N \gg E$) and $\MO(NE)$, otherwise.
    \item In line $10$ follows the greedy search for the \glspl{ev}. Details for Algorithm~\ref{alg:greedy} follow in the next paragraph.
    \item Update $\Bp$ according to the new \glspl{ev} (line \numrange{11}{15}). 
    If the two conditions $N > E$ and $E > (N - E)$ hold, it is more efficient to skip line $11$, \ie, not to reset $\Bp$. 
    Then we can use the modified $\Bp$ of Algorithm~\ref{alg:greedy} and incrementally subtract and remove non-\gls{ev} samples similar as in the regularization in Algorithm~\ref{alg:greedy}.
    This has a time complexity of $\MO((N - E) \cdot E) \Rightarrow \MO(NE)$, since we only need to update the elements in $\Bp$ that are part of $\Theta_E^{t+1}$.
\end{enumerate}

The greedy iteration algorithm is depicted in Algorithm~\ref{alg:greedy} and requires the summations $\Bp$, the combined model $\Theta$, and the budget $K$.
The amount of iterations is limited by $K$ (line~$3$).
In line~4 we take the sample with the highest sum of inclusion probabilities and store it in the \gls{ev} model (line~$5$).
Then follows the bilateral coverage regularization by removing the probability of inclusion of the selected \gls{ev} from the other samples (line \numrange{6}{8}).
In line \numrange{9}{10}, we remove the \gls{ev} from $\Bp$ and $\Theta$.
In the end, we receive the \gls{evm} model $\Theta_E$ containing only the \glspl{ev}.
Note for the mentioned special case in the previous step~$4$, we also need to return the modified $\Bp$ and $\Theta$.

The total asymptotic runtime of the proposed weighted maximum $K$-set cover algorithm is $\MO(N^2)$.
It does not depend on a bisection search as the set cover of Rudd \etal~\cite{rudd2017evm} that has a complexity of $\MO(\log(\epsilon^{-1}) N^2)$, with termination tolerance $\epsilon$.
\renewcommand{\thefigure}{2}
\begin{figure}[!htb]
\renewcommand\figurename{Alg.}  
\hrule\vspace{2pt}
\begin{algorithmic}[1]
\Function{reduce}{$\Bp^t$, $\Theta_E^t$, $\Theta_N^{t+1}$, $K$}
    \LeftComment{Update EV sums: $\MO(EN)$}
    \For{$e$ \textbf{in} $E$}
        \State $\Bp[e] \gets \Bp[e] + \sum_{n \in N} \Psi_e(\Bx_n)$
    \EndFor
    
    \LeftComment{Compute sums for new samples: $\MO(N^2)$}
    \State $\Theta \gets \Theta_E \cap \Theta_N$
    \For{$n$ \textbf{in} $N$}
        \State $p \gets \sum_{i \in |\Theta|} \Psi_n(\Bx_i)$
        \State $\Bp \text{.insert}(p)$
    \EndFor
    
    \State $\Theta_E^{t+1} \gets \text{Greedy}(\Bp, \Theta, K)$
    
    \LeftComment{Compute new $\Bp$: $\MO(E^2)$}
    \State $\Bp^{t+1} \gets \emptyset$
    \For{$i$ \textbf{in} $|\Theta_E^{t+1}|$}
        \State $p \gets \sum_{j \in |\Theta_E^{t+1}|} \Psi_i(\Bx_j)$
        \State $\Bp^{t+1} \text{.insert}(p)$
    \EndFor
\State \Return $\Theta_E^{t+1}$, $\Bp^{t+1}$
\EndFunction
\end{algorithmic}
\vspace{2pt}\hrule\vspace{2pt}
\caption{Detailed version of the proposed class-wise weighted maximum $K$-\setcover model reduction for the \acrfull{evm}.}\label{alg:gmc2}
\end{figure}%
\renewcommand{\thefigure}{3}
\begin{figure}[!htb]
\renewcommand\figurename{Alg.}  
\hrule\vspace{2pt}
\begin{algorithmic}[1]
\Function{greedy}{$\Bp$, $\Theta$, $K$}
    \State $\Theta_E \gets \emptyset$
    \For{$k = 1 \text{ \textbf{to} } K$} \Comment{$\MO(KN)$}
        \State idx $\gets \argmax \Bp$
        \State $\Theta_E^{t+1} \! \text{.insert}(\theta_{\text{idx}})$
        
        \LeftComment{Bilateral coverage regularization: $\MO(N)$}
        \For{$i = 1 \text{ \textbf{to} } |\Theta|$}
            \State $\Bp[i] \gets \Bp[i] - \Psi_i(\Bx_\text{idx})$
        \EndFor
        
        \State $\Bp \text{.remove}(p_\text{idx})$
        \State $\Theta \text{.remove}(\theta_{\text{idx}})$
    \EndFor
\State \Return $\Theta_E$
\EndFunction
\end{algorithmic}
\vspace{2pt}\hrule\vspace{2pt}
\caption{Greedy iterations with bilateral coverage regularization to solve the weighted maximum $K$-\setcover model reduction.}\label{alg:greedy}
\end{figure}%

\renewcommand{\thefigure}{6}
\begin{figure*}[tb]
    \centering
    \vspace{1ex}
    \tikzsetnextfilename{protocol1-cifar-params}
    \includegraphics[width=.99\linewidth]{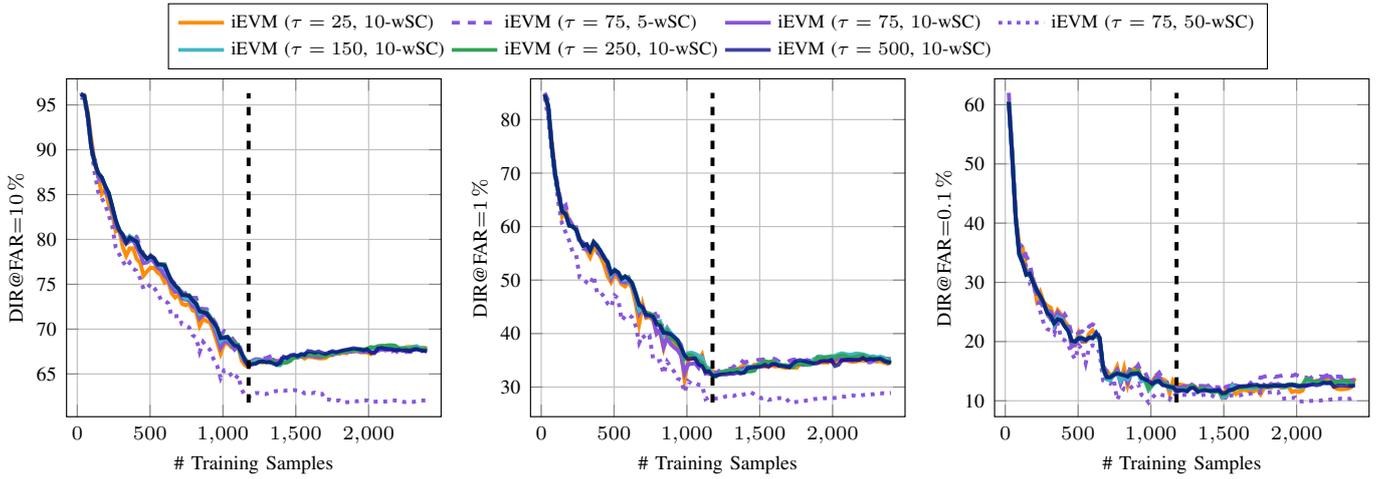}
    \vspace{-1ex}%
    \caption{Different parameterizations of our \acrfull{ievm}. Averaged results over $3$ runs of \protocolA and \cifar{100}. The vertical dashed line determines the batch at which the openness remains constant.}
	\label{fig:cifar100-ievm-params}%
\end{figure*}%
\renewcommand{\thefigure}{7}
\begin{figure*}[tb]
    \centering
    \tikzsetnextfilename{protocol2-icdar-ievm-params}
    \includegraphics[width=.99\linewidth]{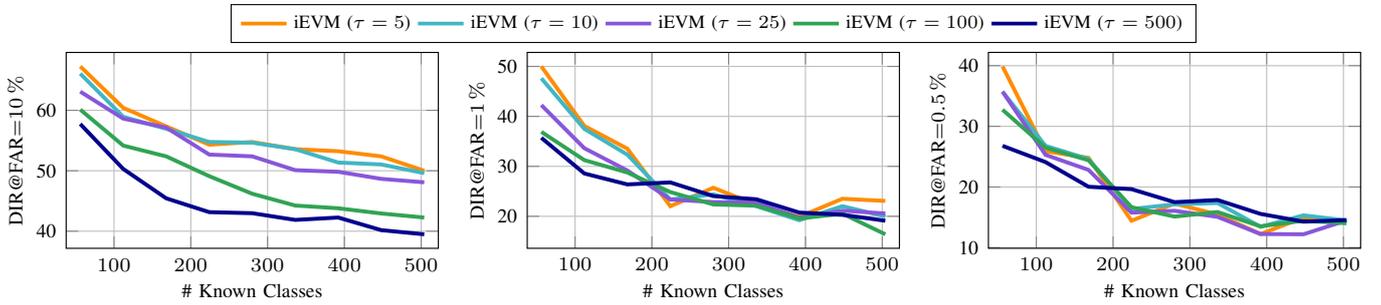}
    \vspace{-1ex}%
    \caption{Different tail size $\tau$ parameterizations of our \acrfull{ievm}. Averaged results over $3$ runs of \protocolB and \icdar.}
	\label{fig:icdar-ievm-params}%
\end{figure*}%

\subsection{Additional Experiments}
In this section we present further experiments of the evaluation with \protocolA and \cifar{100}.
Furthermore, we evaluated the writer identification dataset \icdar~\cite{fiel2017icdar2017} with \protocolB.

\subsubsection{\protocolA\ -- \cifar{100}}
In the main text, we show the result of the \gls{ievm} on \protocolA and \cifar{100} with parameters $\tau = 75$ and the reduction to $K=10$.
Here, we want to present further parameterizations in \cref{fig:cifar100-ievm-params}. As in the main text, the left, middle, and right plots show the \gls{dir} at \glspl{far} of \SI{10}{\percent}, \SI{1}{\percent}, and \SI{0.1}{\percent}.

When comparing the accuracies for different values of $\tau$ at identical $K$, it turns out that the \tailsize $\tau$ has almost no influence on the models' accuracy. 
This is similar to what Günther \etal~\cite{gunther2017opensetface} reported on the \gls{lfw} dataset.

A larger value of $K$ may lead to worse results, as can be seen in the case of iEVM ($\tau = 75$, $50$-wSC). 
This may be counter-intuitive at first glance, considering that classification should perform better with more data.
However, storing more data implies less plasticity and more stability which can interfere with the incremental training adaptability.

\subsubsection{\protocolB\ -- \icdar}
Another \gls{owr} task is writer identification.
Here, we apply \protocolB to the dataset \icdar~\cite{fiel2017icdar2017}.
It contains handwritten pages from the $13^\text{th}$ to $20^\text{th}$ century.
Since the feature extraction is trained on the training set of \icdar, the subsequent classification training and evaluation on the same set would be biased.
Therefore, we take only the test set into account with $5$ pages for each of the $720$ writers.
\SI{30}{\percent} of the classes are selected as unknowns and left in the test split.
For each of the known classes, we leave $1$ sample in the test split, \ie, the training split has $4$ samples for each of the $504$ known classes.
The knowns are split into $9$ batches with $56$ classes and trained incrementally.
This protocol implements an openness from \SIrange{62}{9.3}{\percent}.
The results are averaged over $3$ protocol repetitions.

\paragraph{Implementation Details}
The feature set consists of the \num{6400}-dimensional activation of the penultimate layer of a ResNet20.
It was trained in a self-supervised fashion~\cite{Christlein17ICDAR}. 
The training uses SIFT descriptors that are calculated on patches of $32\times32$ pixels at SIFT keypoints.
The SIFT descriptors are clustered using $k$-means.
Then, the ResNet20 is trained using cross-entropy loss where the patches are used as input and the targets are the cluster center IDs of the patches.
\renewcommand{\thefigure}{8}
\begin{figure*}[tb]
    \centering
    \vspace{1ex}
    \tikzsetnextfilename{protocol2-icdar2}
    \includegraphics[width=.99\linewidth]{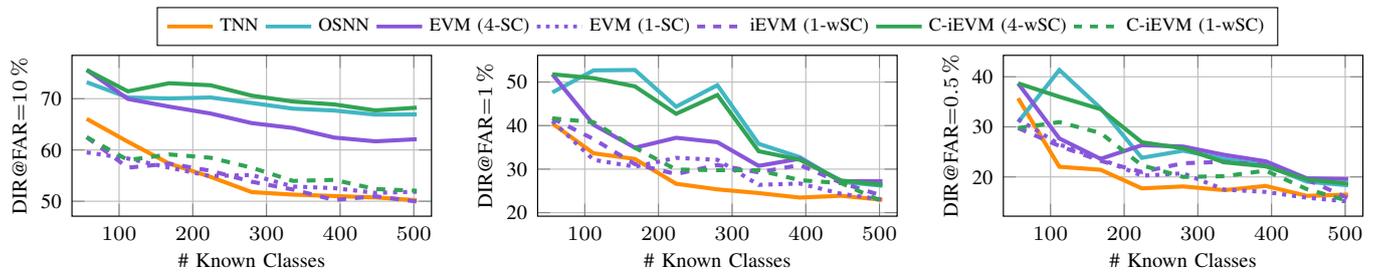}
    \vspace{-1ex}%
    \caption{Averaged results over $3$ runs of \protocolB on \icdar. Set cover and our weighted maximum $K$-set cover reduction to $K$ \acrfullpl{ev} are denoted as $K$-SC and $K$-wSC, respectively.}
	\label{fig:icdar}%
\end{figure*}%

\paragraph{Hyperparameter Evaluation}
The experiments on \cifar{100} and \protocolA show, similar as the previous work of G{\"u}nther \etal~\cite{gunther2017opensetface}, that the \tailsize parameter $\tau$ has only a minor impact on the results.
However, we noticed that this does not apply to \protocolB and \icdar as visualized in \cref{fig:icdar-ievm-params}.
The experiments show that a small \tailsize ($\tau \in \{5, 10\}$) achieves a better \gls{dir} at a high \gls{far} of \SI{10}{\percent}.
This difference degrades over the class-wise increments at medium and small \glspl{far}  of \SI{1}{\percent} and \SI{0.5}{\percent}.
Rudd \etal~\cite{rudd2017evm} state that a larger \tailsize leads to higher coverage. 
This implies that for \icdar a high coverage and little \openspace is less favorable and a steep decision boundary is beneficial.

\paragraph{Results}
The comparison to the other baseline methods follows in \cref{fig:icdar}.
All \glspl{evm} use a \tailsize $\tau = 5$.
The C-iEVM without model reduction performs comparable to the \gls{osnn} and both outperform the conventional \gls{evm}.
The boundary case of a model reduction to a single \gls{ev} per class does not lead to an improvement in this evaluation.
In contrast to this result, we note that the evaluation of Protocol I on \cifar{100} performed much better with model reduction.
However, the representation of a class via a single sample is challenging and heavily depends on the class distribution.

\end{document}